\documentclass[11pt,letterpaper]{article}
\usepackage{times}
\usepackage{latexsym,bm}
\usepackage{amsmath,amsfonts}
\usepackage{multirow}
\usepackage{wrapfig}
\usepackage{url}
\usepackage{appendix}

\usepackage{iclr2019_conference,times}

\usepackage{graphicx,subcaption}
\RequirePackage[textsize=scriptsize]{todonotes}
\DeclareMathOperator*{\argmax}{arg\,max}

\newcommand{\vek}[1]{{\bf {#1}}}
\newcommand{\vx}{{\vek{x}}}
\newcommand{\vs}{{\vek{s}}}
\newcommand{\vy}{{\vek{y}}}
\newcommand{\cx}{{\vek{H}}}
\newcommand{\valpha}{{\bm{\alpha}}}

\newcommand{\vw}{{\vek{w}}}
\newcommand{\vh}{{\vek{h}}}

\newcommand{\cH}{{{\cal H}}}
\newcommand{\vyp}{{\vek{\hat{y}}}}
\newcommand{\bleu}{{\text{\sc bleu}}}
\long\def\ignore#1{}

\ignore{
   Real work:
   Calibration of FairSeq model?
   Since MMCE is published, should we apply it for seq2seq learning.
   Check if RNMT code is available in ternsor2tensor package
   Kitchen sink ben recht nips 2008 for performance

   Harness a second T2T model
   
}

\iclrfinalcopy

\title{Calibration of Encoder Decoder Models for Neural Machine Translation}

\author{Aviral Kumar\thanks{Work done when the author was a student at IIT Bombay} \\
  University of California Berkeley\\
  {\tt aviralk@berkeley.edu} \\\And
  Sunita Sarawagi \\
  Indian Institute of Technology Bombay\\
  {\tt sunita@iitb.ac.in} \\}

\date{}

\begin{document}
\maketitle


\begin{abstract}
We study the calibration of several state of the art neural machine translation (NMT) systems built on attention-based encoder-decoder models. For structured outputs like in NMT, 
calibration is important not just for reliable confidence with predictions, but also for proper functioning of beam-search inference.  We show that most modern NMT models are surprisingly miscalibrated even when conditioned on the true previous tokens.  Our investigation leads to two main reasons --- severe miscalibration of EOS (end of sequence marker) and suppression of attention uncertainty.  We design recalibration methods based on these signals and demonstrate improved accuracy, better sequence-level calibration, 
and more intuitive results from beam-search.
\end{abstract}

\section{Introduction}
Calibration of supervised learning models is a topic of continued interest in machine learning and statistics~\cite{05predicting,challenge,Crowson16,GuoPSW17}. 
Calibration requires that the probability a model  assigns to a prediction equals the true chance of correctness of the prediction. For example, if a calibrated model $M$ makes 1000 predictions with probability values around $0.99$, we expect 990 of these to be correct. If $M$ makes another 100 predictions with probability 0.8, we expect around 80 of these to be correct.  
Calibration is important in real-life deployments of a model since it ensures interpretable probabilities, and plays a crucial role in reducing prediction bias~\cite{PleissRWKW17}.  In this paper we show that for structured prediction models calibration is also important for sound working of the inference algorithm that generates structured outputs.

Much recent work have studied calibration of modern neural networks for scalar predictions~\cite{GuoPSW17,LakshmiN17,HendrycksG17,louizos17a,PereyraTCKH17,kumar2018,Kuleshov18}. A surprising outcome is that modern neural networks have been found to be miscalibrated in the direction of over-confidence, in spite of a statistically sound log-likelihood based training objective.

We investigate calibration of attention-based encoder-decoder models for sequence to sequence (seq2seq) learning as applied to  neural machine translation. 
We measure calibration of token probabilities of three modern neural architectures for translation --- NMT~\cite{nmt}, GNMT~\cite{Wu16}, and the Transformer model~\cite{Vaswani2017} on six different benchmarks.  We find the output token probabilities of these models to be poorly calibrated.
This is surprising because the  output distribution is conditioned on {\em true} previous tokens (teacher forcing) where there is no train-test mismatch unlike when we condition on {\em predicted} tokens where there is a risk of exposure bias~\cite{Bengio2015,marc2016sequence,rewardAugmented,beamsearch}.
We show that such lack of calibration can explain the counter-intuitive 
\bleu\ drop with increasing beam-size~\cite{KoehnK17}.


We dig into root causes for the lack of calibration and pin point two primary causes: poor calibration of the EOS token and attention uncertainty.  
Instead of generic temperature based fixes as in~\cite{GuoPSW17}, 
we propose a parametric model to recalibrate as a function of input coverage, attention uncertainty, and token probability. 
We show that our approach leads to improved token-level calibration. 
We demonstrate three advantages of a better calibrated model.
First, we show that the calibrated model better correlates probability with \bleu\ and that leads to \bleu\ increment by up to 0.4 points just by recalibrating a pre-trained model. 
Second, we show that the calibrated model has better calibration on the per-sequence \bleu metric, which we refer to as sequence-level calibration and was achieved just by recalibrating token-level probabilities. 
Third, we show that improved calibration diminishes the drop in \bleu\ with increasing beam-size. Unlike patches like coverage and length penalty~\cite{Wu16,He2016ImprovedNM,Yang2018BreakingTB}, inference on calibrated models also yields reliable probabilities.

%
 


\section{Background and Motivation}
\label{sec:measure}
We review the model underlying modern NMT systems\footnote{see \cite{koehn17} for a more detailed review}, then discuss measures for calibration.
\subsection{Attention-based NMT}
State of the art NMT systems use an attention-based encoder-decoder neural network for modeling $\Pr(\vy | \vx,\theta)$ 
over the space of discrete output translations of an input sentence $\vx$ where $\theta$ denotes the network parameters.  Let $y_1,\ldots,y_n$ denote the tokens in a sequence $\vy$ and $x_1,\ldots,x_k$ denote tokens in $\vx$.  Let $V$ denote output vocabulary.
A special token EOS $\in V$
marks the end of a sequence in both $\vx$ and $\vy$. 
First, an encoder (e.g. a bidirectional LSTM) transforms each $x_1,\ldots,x_k$ 
into real-vectors $\vh_1,\ldots,\vh_k$.  
%
%
The Encoder-Decoder (ED) network factorizes $\Pr(\vy | \vx,\theta)$ as
\begin{equation}
\label{eq:chain}
\Pr(\vy | \vx,\theta)=\prod_{t=1}^n\Pr(y_t |\vy_{< t},\vx,\theta)
\end{equation}
where $\vy_{< t}=y_1,\ldots,y_{t-1}$. The decoder computes each $\Pr(y_t | \vy_{< t},\vx,\theta)$ as
\begin{equation}
\Pr(y_t | \vy_{< t},\vx,\theta) = \text{softmax}(\theta_{y_t} F_\theta(\vs_t, \cx_t))
\end{equation}
where $\vs_t$ is a decoder state summarizing $y_1,\ldots y_{t-1}$; $\cx_t$ is attention weighted input:
\begin{equation}
\label{eq:attn}
\cx_t = \sum_{j=1}^k \vh_j \alpha_{jt},~\valpha_{t} = \text{softmax}(A_\theta(\vh_j,\vs_t))
\end{equation}
$A_\theta(.,.)$ is the attention unit.

\noindent
During training given a $D=\{(\vx_i,\vy_i)\}$, we find $\theta$ to minimize negative log likelihood~(NLL):
\begin{equation}
\label{eq:nll}
\text{NLL}(\theta) = 
-\sum_{i\in D}\sum_{t=1}^{|\vy_i|} \log\Pr(y_{it} | \vy_{i,< t},\vx_i,\theta)
\end{equation}
During inference given a $\vx$, we need to find the $\vy$  that maximizes $\Pr(\vy|\vx)$. This is intractable given the full dependency (Eq:~\ref{eq:chain}). Approximations like beam search with a beam-width parameter $B$ (typically between 4 and 12) maintains $B$ highest probability prefixes which are grown token at a time.  At each step beam search finds the top-B highest probability tokens from $\Pr(y | \vy_{b,< t},\vx,\theta)$ for each prefix $\vy_{b,<t}$ until a EOS is encountered. 

\subsection{Calibration: Definition and Measures}
Our goal is to study, analyze, and fix the calibration of the next token distribution $Pr(y_{it} | \vy_{i,< t},\vx_i,\theta)$ that is used at each inference step. We first define calibration and how it is measured.  Then, we motivate the importance of calibration in beam-search like inference for sequence prediction.  

We use the short-form $P_{it}(y)$ for $Pr(y_{it} | \vy_{i,< t},\vx_i,\theta).$
A prediction model $P_{it}(y)$ is well-calibrated if for any value $\beta \in [0,1]$, of all predictions $y \in V$ with probability $\beta$, the fraction correct is $\beta$. That is, the model assigned probability represents the chance of correctness of the prediction.  

Calibration error measures the mismatch  between the model assigned probability (also called confidence) and fraction correct.  To measure such mismatch on finite test data we bin the range of $\beta$ [0,1] into equal sized bins $I_1,\ldots,I_M$ (e.g. [0,0.1),[0.1,0.2), etc) and sum up the mismatch in each bin.  Say, we are given a test data with $L$  $P_{it}(y)$ distributions spanning different sequence $i$ and step $t$ combinations.  Let $y_{it}$ denote the correct output and $\hat{y}_{it}=\argmax_y P_{it}(y)$ denote the prediction; its prediction confidence is then $c_{it} = P_{it}(\hat{y}_{it})$. 
Within each bin $I_b$, let $S_b$ denote all $(i,t)$ where confidence $c_{it}$ value falls within that bin. Over each $S_b$ we measure, (1) the fraction correct or accuracy $A_b$, that is the fraction of cases in $S_b$ where  $y_{it}=\hat{y}_{it}$, (2) the average $c_{it}$ value, called the average confidence $C_b$, (3) the total mass on the bin $w_b=$ the fraction of the $L$ cases in that bin.  A graphical way to measure 
calibration error is via reliability plots that shows average confidence $C_b$ on the x-axis against average accuracy $A_b$.  In a well-calibrated model where confidence matches with correctness, the plot lies on the diagonal. Figure~\ref{fig:overall} shows several examples of calibration plots of two models with $M=20$ bins each of size 0.05. The bins have been smoothed over in these figures.  The absolute difference between the diagonal and the observed plot scaled by bin weight $w_b$ is called {\bf expected calibration error (ECE)}. 
ECE considers only the highest scoring prediction from each $P_{it}(y)$ but since beam-search reasons over probability of multiple high scoring tokens, we extended and used a weighted version of ECE that measures calibration of the \emph{entire distribution}.
We describe ECE and weighted ECE more formally below, and also provide an example to motivate the use our weighted ECE metric for structured prediction tasks.

\subsubsection{Expected Calibration Error (ECE)}
\label{sec:appendix_ece_weighted_ece}
ECE is defined when a model makes a single prediction $\hat{y}$ with a confidence $p$.  In the case of scalar prediction or considering just the topmost token in structured prediction tasks, the prediction is $\hat{y}_{it}=\argmax_y P_{it}(y)$ with $P_{it}(\hat{y}_{it})$ as confidence.  Let $C_{it}(y)=\delta(y_{it} = y)$ denote if $y$ matches the correct label $y_{it}$ at $(i,t)$. 

First partition the confidence interval [0..1] into $M$ equal bins $I_1,\ldots,I_M$.  Then in each bin measure the absolute difference between the accuracy and confidence of predictions in that bin.  This gives the expected calibration error (ECE) as:
\begin{equation}
\frac{1}{L}\sum_{b=1}^M\left|\sum_{i,t:P_{it}(\hat{y}_{it}) \in I_b} C_{it}(\hat{y}_{it})-P_{it}(\hat{y}_{it})\right|
\end{equation}
where $L=\sum_i^N |\vy_i|$ is total output token lengths (or total number of scalar predictions made).
Since beam-search reasons over probability of multiple high scoring tokens, we wish to calibrate the entire distribution. If V is the vocabulary size, we care to calibrate all $LV$ predicted probabilities. A straightforward use of ECE that treats these as $LV$ independent scalar predictions is incorrect, and is not informative.

\subsubsection{Weighted Expected Calibration Error (Weighted ECE)}
Weighted ECE is given by the following formula: (various symbols have usual meanings as used in the rest of this paper)
$$\frac{1}{L}\sum_{b=1}^M\left|\sum_{i,t}\sum_{y:P_{it}(y) \in I_b} P_{it}(y)(\delta(y_{it} = y)-P_{it}(y))\right|.$$ 
We motivate our definition as applying ECE on a classifier that  predicts label ${y}$ with probability proportional to its confidence $P_{it}(y)$ instead of the highest scoring label deterministically.

\paragraph{Example} This example highlights how weighted ECE  calibrates the full distribution.
Consider two distributions on a V of size 3: $P_{1}(.)=[0.4,~0.1,~0.5]$ and $P_{2}(.)=[0.0,~0.5,~0.5]$. For both let the first label be correct.  Clearly, $P_1$ with correct label probability of 0.4 is better calibrated than $P_2$.  But ECE of both is the same at $|0-0.5|=0.5$ since both of theirs highest scoring prediction (label 3) is incorrect.   In contrast, with bins of size 0.1, weighted ECE will be $0.4|1-0.4|+0.1|0-0.1|+0.5|0-0.5|=0.42$ for $P_{1}$ which is less than 0.5 for $P_{2}$.  
Such fine-grained distinction is important for beam-search and any other structured search algorithms with large search spaces. In the paper we used ECE to denote weighted ECE.

\noindent

\subsection{Importance of Calibration}
In scalar classification models, calibration as a goal distinct from accuracy maximization, is motivated primarily by interpretability of the confidence scores.  In fact, the widely adopted fix for miscalibration, called temperature scaling, that scales the entire $P_{it}(y)$ distribution by a constant temperature parameters $T$ as $\propto P_{it}(y)^{\frac{1}{T}}$.  does not change the relative ordering of the probability of the $y$, and thus leaves the classification accuracy unchanged. For sequence prediction models, we show that calibration of the token distribution is important also for the sound working of the beam-search inference algorithm.  
%
Consider an example: say we have an input sequence for which the correct two-token sequence is "{\em That's awesome}". Let's say the model outputs a miscalibrated distribution for the first token position:
$$P_{1}(y=\mbox{\em It's})=0.4, P_{1}(y=\mbox{\em That's})=0.6$$ where the ideal model should have been $$P^*_{1}(y=\mbox{\em It's})=0.3, P^*_{1}(y=\mbox{\em That's})=0.7.$$ Assume at $t=2$, the model is calibrated and $$P_{2}(\mbox{\em ok}|\mbox{\em It's})=0.91, P_{2}(\mbox{\em awesome}|\mbox{\em That's})=0.6.$$  The highest probability prediction from the uncalibrated model is {\em It's ok} with probability $0.4\times 0.91$, whereas from the calibrated model is {\em That's awesome}. Thus, accuracy of the model $P$ is 0 and the calibrated $P^*$ is 1 even though the relative ordering of token probabilities is the same in $P$ and $P^*$.  This example also shows that if we used beam size =1, we would get the correct prediction although with the lower score $0.6\times 0.6$, whereas the higher scoring ( $0.4\times 0.91$) prediction obtained at beam size=2 is wrong. 

More generally, increasing the beam size almost always outputs a higher scoring prefix but if the score is not calibrated that does not guarantee more correct outputs. The more prefixes we have with over-confident (miscalibrated) scores, the higher is our chance of over-shadowing the true best prefix in the next step, causing accuracy to drop with increasing beam-size.  This observation is validated on real data too where we observe that improving calibration captures the BLEU drop with increasing beam size.

\section{Calibration of existing models}
\label{sec:models}
We next study the calibration of six state-of-the-art publicly available pre-trained  NMT models  on various WMT+IWSLT benchmarks.
The first five are from Tensorflow's NMT codebase~\cite{luong17} \footnote{https://github.com/tensorflow/nmt\#benchmarks}: En-De GNMT (4 layers), En-De GNMT (8 layers), De-En GNMT, De-En NMT, En-Vi NMT. They all use multi-layered LSTMs arranged either in the GNMT architecture~\cite{Wu16} or standard NMT architecture~\cite{nmt}.
The sixth En-De T2T, is the pre-trained Transformer model\footnote{https://github.com/tensorflow/tensor2tensor/tree/master/\\tensor2tensor, pre-trained model at  https://goo.gl/wkHexj}. (We use T2T and Transformer interchangeably.)
The T2T replaces LSTMs with self-attention~\cite{Vaswani2017} and uses multiple attention heads, each with its own attention vector. 

\begin{figure}[h]
\begin{center}
\includegraphics[scale=0.5]{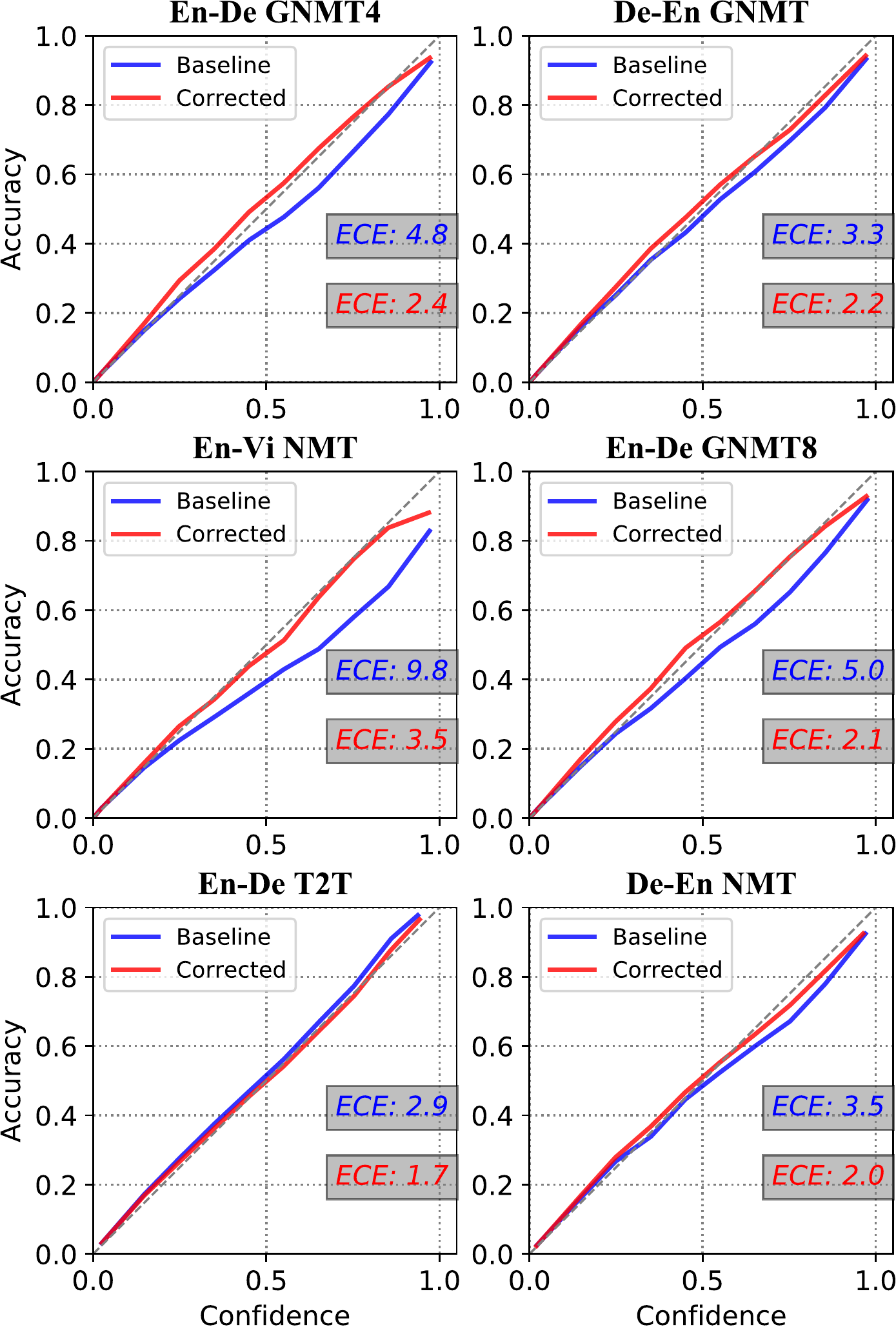}
    \caption{\label{fig:overall} \small Reliability Plots for various baseline models on the test sets along with their ECE values(Blue). The x-axis is expected confidence after binning into 0.05 sized bins and y-axis is accuracy in that confidence bin. 
    Reliability plots for calibrated (corrected) models (Red). ECE values in corresponding colors. 
    Test sets are mentioned in the corresponding references.}
\end{center}
\end{figure}

Figure~\ref{fig:overall} shows calibration as a reliability plot where x-axis is average weighted confidence and y-axis is average weighted accuracy. The blue lines are for the original models and the red lines are after our fixes (to be ignored in this section).   The figure also shows calibration error (ECE).   We observe that all six models are miscalibrated to various degrees with ECE ranging from 2.9 to 9.8.  For example,  in the last bin of En-Vi NMT the average predicted confidence is 0.98 whereas its true accuracy is only 0.82.
Five of the six models are overly confident.  
The transformer model attempts to fix the over-confidence by using a soft cross-entropy loss 
that assigns a probability $t_y= 0.9$ to the correct label and probability $t_y=\frac{0.1}{V-1}$ to all others as follows:
\vspace{-1em}
\begin{equation}
\label{eq:smoothNLL}
\min_\theta 
- \sum_{i=1}^N\sum_{t=1}^{|\vy_i|} \sum_{y}t_y\log\Pr(y| \vy_{i,< t},\vx_i,\theta) 
\end{equation}
With this loss, the over-confidence changes to slight under-confidence. While an improvement over the pure NLL training, we will show how to enhance its calibration even further.



This observed miscalibration was surprising given the tokens are conditioned on the {\em true} previous tokens (teacher forcing). We were expecting biases when conditioning on {\em predicted} previous tokens because that leads to what is called as "exposure bias"~\cite{Bengio2015,marc2016sequence,rewardAugmented,beamsearch}. In teacher-forcing, the test scenario matches the training scenario where the 
NLL training objective~(Eq~\ref{eq:nll}) is statistically sound --- it is minimized when the fitted distribution matches the true data distribution~\cite{hastie01statisticallearning}.

\subsection{Reasons for Miscalibration}
\label{sec:reasons}
In this section we seek out reasons for the observed miscalibration of modern NMT models.
For scalar classification~\citet{GuoPSW17} discusses reasons for poor calibration of modern neural networks (NN). A primary reason  is  that the high capacity of NN causes the negative log likelihood (NLL) to overfit without overfitting 0/1 error~\cite{DBLP:journals/corr/ZhangBHRV16}. 
%
We show that for sequence to sequence learning models based on attention and with large vocabulary, a different set of reasons come into play.  We identify three of these.  While these are not exclusive reasons, we show that correcting them improves calibration and partly fixes other symptoms of miscalibrated models. 
\begin{figure*}[h]
\begin{center}
\begin{subfigure}[b]{1.0\textwidth}
\begin{center}
\includegraphics[scale=0.4]{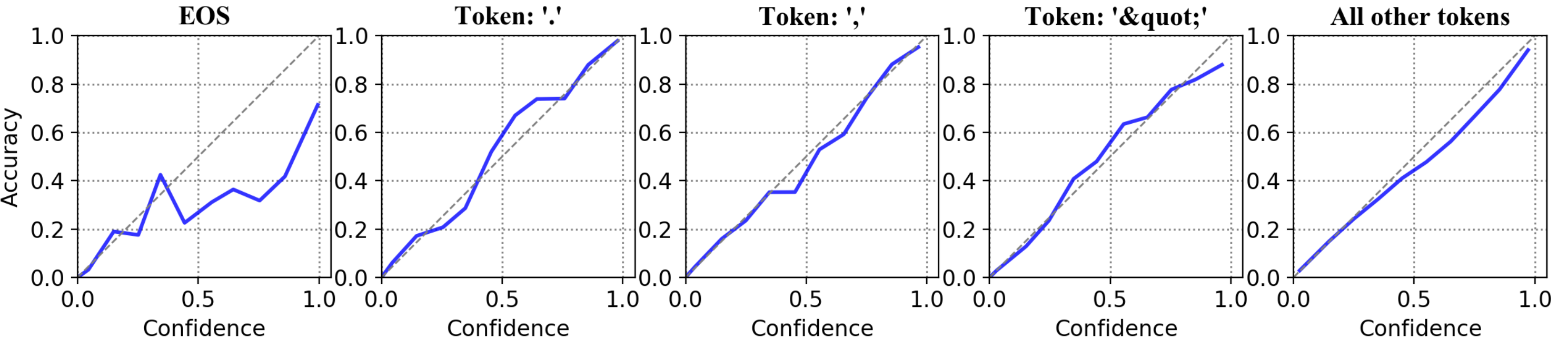}
\end{center}
\caption{En-De GNMT(4)}
\end{subfigure}
\begin{subfigure}[b]{1.0\textwidth}
\begin{center}
\includegraphics[scale=0.4]{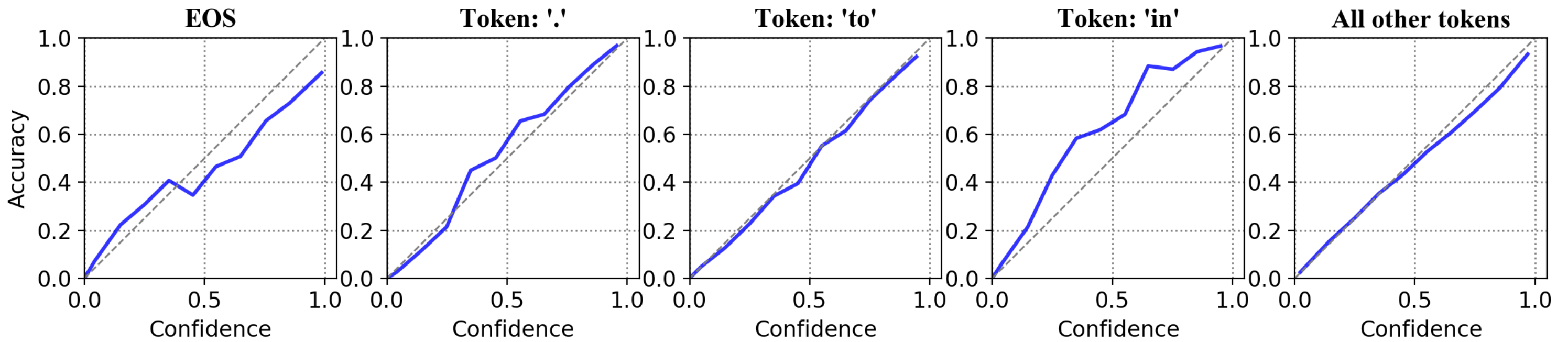}
\end{center}
\caption{De-En GNMT(4)}
\end{subfigure}
\begin{subfigure}[b]{1.0\textwidth}
\begin{center}
\includegraphics[scale=0.4]{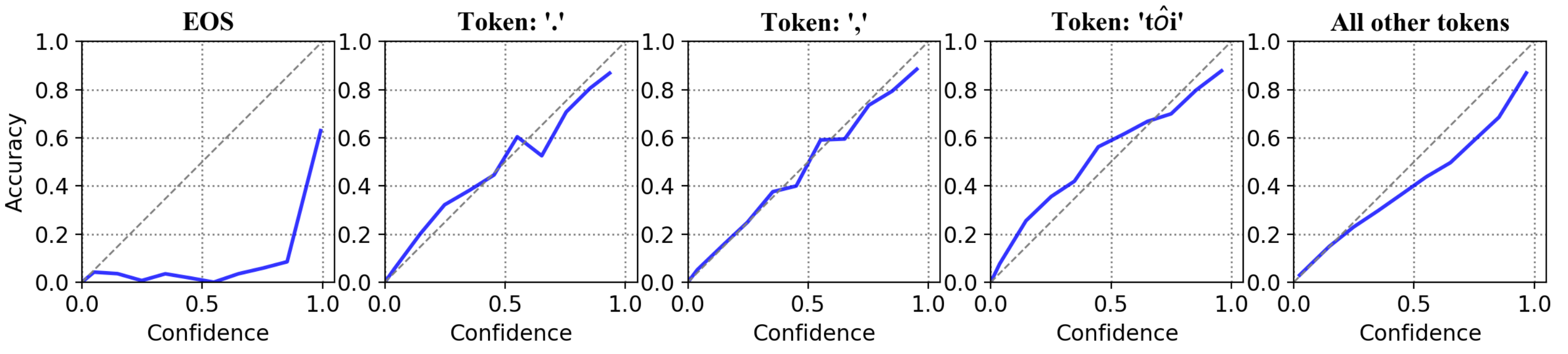}
\end{center}
\caption{En-Vi NMT}
\end{subfigure}
\begin{subfigure}[b]{1.0\textwidth}
\begin{center}
\includegraphics[scale=0.4
]{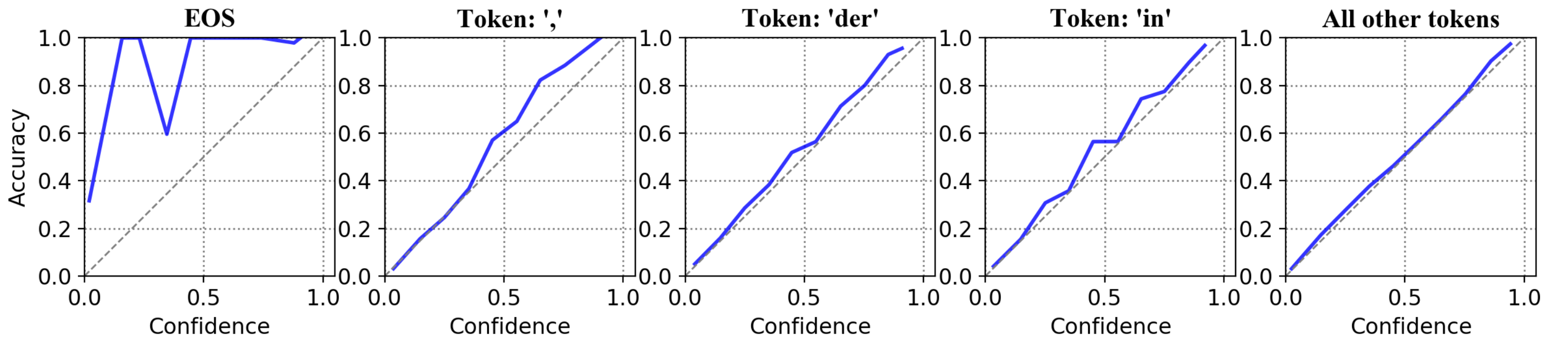}
\end{center}
\caption{En-De T2T (Transformer)}
\end{subfigure}
\caption{\label{fig:token} \small Tokenwise Calibration plots for some of the models. Note the miscalibration of EOS vs the calibration of other tokens. All other tokens roughly show a similar trend as the overall calibration plot.}
\end{center}
\end{figure*}
\subsection{Poor calibration of EOS token}
To investigate further we   
drill down to token-wise calibration. Figure~\ref{fig:token} shows the plots of EOS, three other frequent tokens, and  the rest for four models. 
Surprisingly, EOS is calibrated very poorly and is much worse than the overall calibration plots in Figure~\ref{fig:overall} and other frequent tokens. For NMT and GNMT models EOS is over-estimated, and for T2T the EOS is under-estimated.  For instance, for the En-De GNMT(4) model (top-row, first column in Fig~\ref{fig:token}), out of all EOS predictions with confidence in the [0.9, 0.95] bin only 60\% are correct.  Perhaps these encoder-decoder style models do not harness enough signals to reliably model the end of a sequence.  One such important signal is coverage of the input sequence. While coverage has been used heuristically in beam-search inference~\cite{Wu16}, we propose  a more holistic fix of the entire distribution using coverage as one of the features in Section~\ref{sec:fix}.



\subsection{Uncertainty of Attention}
We conjectured that a second reason for over-confidence could be the uncertainty of attention.
A well-calibrated model must express all sources of prediction uncertainty in its output distribution. Existing attention models average out the attention uncertainty of $\valpha_{t}$ in the input context $\cx_t$ (Eq:~\ref{eq:attn}).  Thereafter,  $\valpha_{t}$ has no influence on the output distribution.  We had conjectured that this would manifest as worser calibration for high entropy attentions $\valpha_{t}$, and this is what we observed empirically.
In Table~\ref{tab:attn} we show ECE partitioned across whether the entropy of  $\valpha_{t}$ is high or low on five\footnote{We drop the T2T model since measuring attention uncertainty is unclear in the context of multiple attention heads.} models. Observe that ECE is higher for high-entropy attention. 
\begin{table}[ht!]
\begin{center}
 \begin{tabular}{|l| r |r|} 
 \hline
 \textbf{Model Name} & \textbf{Low $\mathcal{H}$} & \textbf{High $\mathcal{H}$} \\ 
 \hline\hline
 En-Vi NMT & 9.0  & 13.0 \\ 
 \hline
 En-De GNMT(4) & 4.5 & 5.3 \\
 \hline
 En-De GNMT(8) & 4.8 & 5.4 \\
 \hline
 De-En GNMT$^{\Box}$ & 3.8 & 2.3 \\ 
 \hline
 De-En GNMT$^{\otimes}$ & 3.9 & 5.9 \\
 \hline
 De-En NMT & 2.3 & 4.1 \\
 \hline
\end{tabular}
\end{center}
\caption{\label{tab:attn} \small ECE(\%) for the high and low attention entropy zones. High entropy is defined as $\mathcal{H} \geq 1.0$; ($\Box$ represents the ECE for the entire set of samples, $\otimes$ represents the ECE for the samples with prediction probability in $0.8-1.0$ -- this was done to see how attention entropy correlates with calibration in the high confidence prediction range). }
\end{table}

\subsection{Head versus Tail tokens}
\begin{center}
\begin{figure}[ht!]
\centering
\begin{center}
\begin{subfigure}{0.99\hsize}
\includegraphics[scale=0.4]{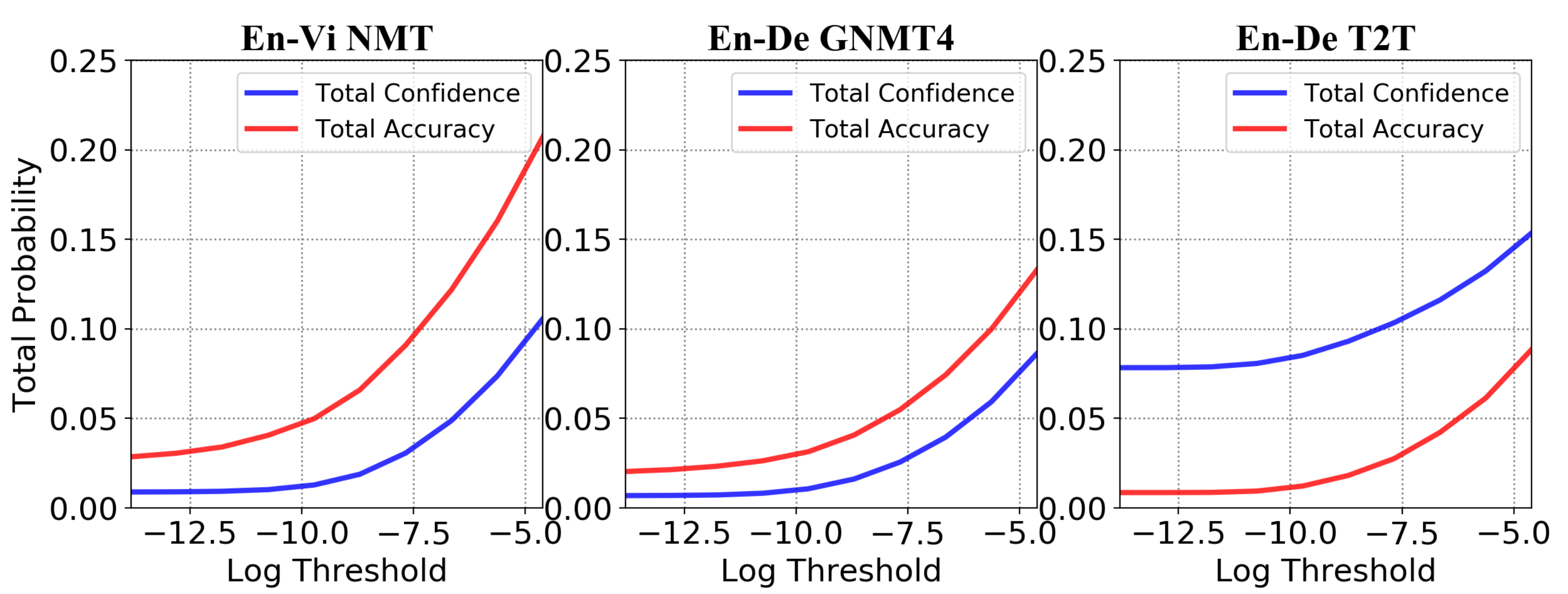}
\caption{\label{fig:tail} \small Tail calibration plots for three models}
\end{subfigure}\\
\begin{subfigure}{0.99\hsize}
\includegraphics[scale=0.4]{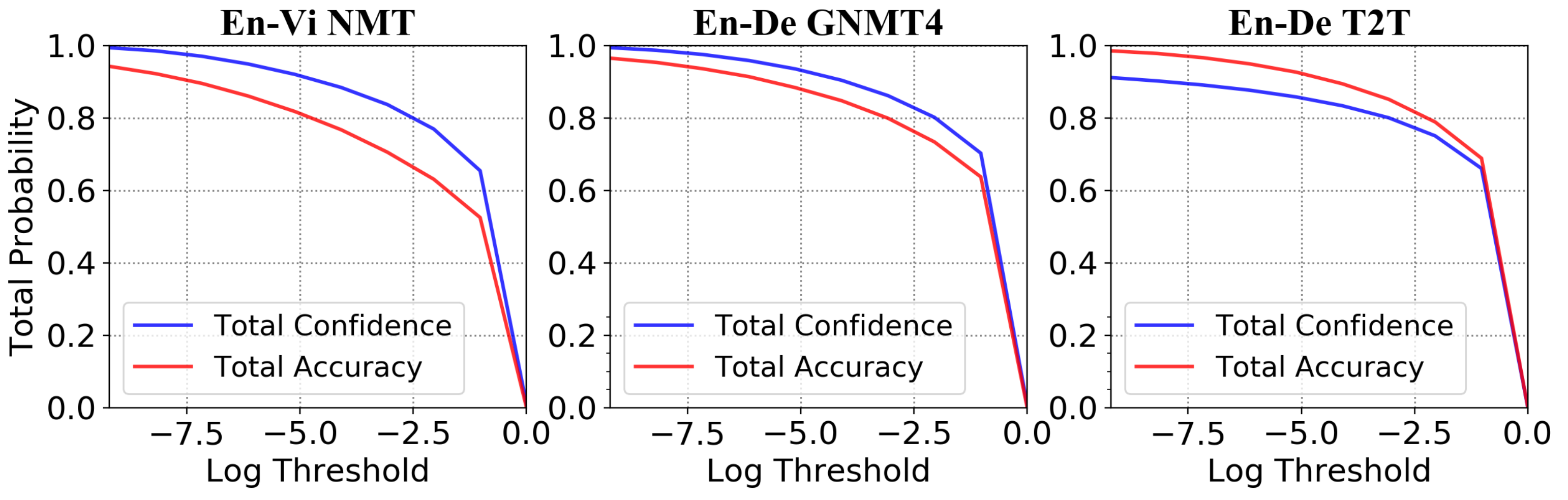}
\caption{\label{fig:head} \small Head calibration plots for three models.}
\end{subfigure}
\end{center}
\caption{\small Tail and Head Calibration Plots for 3 models. Note that the head is overestimated in GNMT/NMT, underestimated in T2T and the tail shows the opposite trend. Here the x-axis corresponds to the log of the fraction of vocabulary that is classified as tail prediction. }
\end{figure}
\end{center}

The large vocabulary and the softmax bottleneck~\cite{yang2018breaking} was another reason we investigated. We studied the calibration for tail predictions (the ones made with low probability) in contrast to the head in a given softmax distribution.  In Figure~\ref{fig:tail} for different thresholds $T$ of log probability (X-axis), we show total true accuracy (red) and total predicted confidence (blue) for all predictions with confidence less than $T$.  In Figure~\ref{fig:head} we show the same for head predictions with confidence $>T$.  The first two from GNMT/NMT under-estimate tail (low) probabilities while over-estimating the head. 
The T2T model shows the opposite trend. This shows that the phenomenon of miscalibration manifests in the entire softmax output and motivates a method of recalibration that is sensitive to the output token probability.

\section{Reducing Calibration Errors}
\label{sec:fix}
For modern neural classifiers \citet{GuoPSW17} compares several post-training fixes 
and finds temperature scaling to provide the best calibration without dropping accuracy.  This method chooses a positive temperature value $T$ and transforms the $P_{it}(y)$ distribution as $\propto P_{it}(y)^{\frac{1}{T}}$. 
  The optimal $T$ is obtained by maximizing NLL on a held-out validation dataset.  

Our investigation in Section~\ref{sec:reasons} showed that calibration of different tokens in different input contexts varies significantly.
We propose an alternative method, where the temperature value is not constant but varies based on the entropy of the attention, the log probability (logit) of the token, the token's identity (EOS or not), and the input coverage. 
%
At the $t$-th decoding step, let $a_t = \cH(\valpha_t)$ denote the entropy of the attention vector $\valpha_t$ and the logit for a token $y$ at step $t$ be $l_{ty}=\log\Pr(y | \vy_{< t},\vx,\theta)$. We measure coverage $c_t$ as the fraction of input tokens with cumulative attention until $t$ greater than a threshold $\delta$. We used $\delta = 0.35$. 
%
Using $(a_t,c_t,l_{ty})$ we compute the (inverse of ) temperature for scaling token $y$ at step $t$ in two steps. We first correct the extreme miscalibration of EOS by learning a correction as a function of the input coverage $c_t$ as follows:
$$ l'_{ty} = l_{ty} + {[[y = \text{eos}]]}\log \big(\sigma(w_1 (c_t - w_2))\big)$$
This term helps to dampen EOS probability when input coverage $c_t$ is low and $w_1,w_2$ are learned parameters.
Next, we correct for overall miscalibration by using a neural network to learn variable temperature values as follows:
$$T^{-1}_{ty}(a_t,l'_{ty}, c_t|\vw) =  g_\vw(a_t) \cdot h_\vw(l'_{ty}) $$
where $g_\vw(.)$ and $h_\vw(.)$ are functions with parameters $\vw$. 
For each of $g_\vw(.)$ and $h_\vw(.)$, we use a 2-layered feed-forward network with hidden ReLu activation, three units per hidden layer, and a sigmoid activation function to output in range $(0, 1)$. 
Since the T2T model under-estimates probability, we found that learning was easier if we added 1 to the sigmoid outputs of $h_\vw$ and $g_\vw$ before multiplying them to compute the temperature.
%
We learn parameters $\vw$ (including $w_1$ and $w_2$) by minimizing NLL on temperature adjusted logits using a validation set $D_V$.
\begin{equation*}
\label{eq:cal}
-\sum_{i \in D_V} \sum_{t=1}^{|\vy_i|} \log (\text{softmax}_y (l'_{ity}T^{-1}_{ty}(a_{it},l'_{ity}, c_{it}|\vw))[y_{it}])
\end{equation*}
where $l_{ity} =  \log\Pr(y | \vy_{i,< t},\vx_i,\theta)$ and $l'_{ty}$ is as defined earlier.
The held-out validation set $D_V$ was created using a 1:1 mixture of 2000 examples sampled from the train and dev set. The dev and test distributions are quite different for WMT+IWSLT datasets.  So we used a mixture of dev and train set for temperature calibration rather than just the dev set for generalizability reasons. Just using the dev set defeats the purpose of calibration, as the temperature-based calibration method can potentially overfit to a particular distribution (dev) whereas using a mixture of dev and train will prevent this overfitting and hence, provide us with a model that is more likely to generalize to a third (test) distribution.  


\section{Experimental Results}
\label{sec:expts}
We first show that our method manages to significantly reduce calibration error on the test set.   
Then we present two outcomes of a better calibrated model:  (1) higher accuracy, and (2) reduced BLEU drop with increasing beam-size.

\ignore{
\begin{figure}[ht!]
\begin{center}
\includegraphics[width=\hsize]{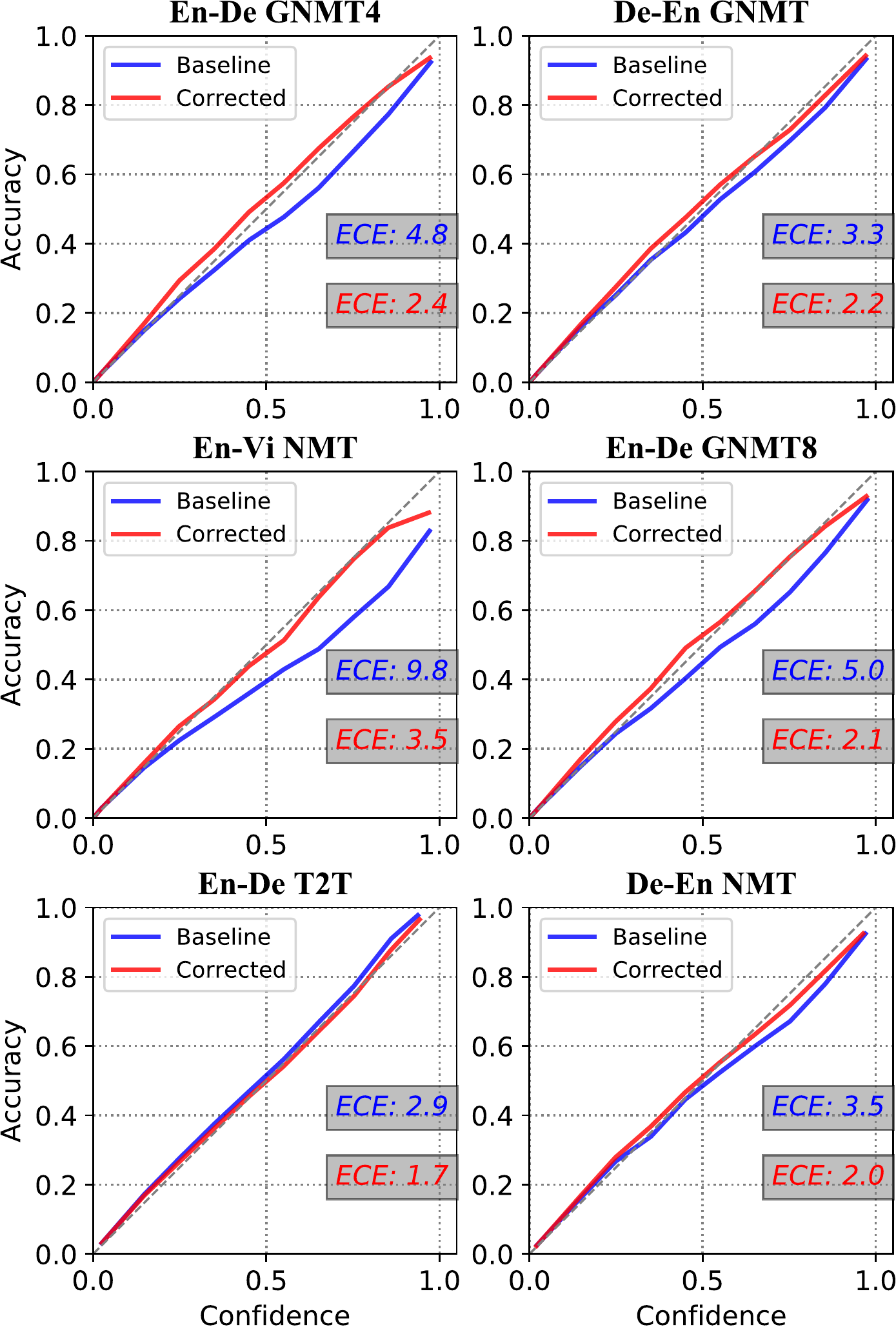}
     \caption{\label{fig:fix}Corrected Overall Calibration Plots for various models on the corresponding test sets.}
 \end{center}
 \end{figure}
 }
\subsection{Reduction in Calibration Error}
Figure~\ref{fig:overall}  shows that our method (shown in red) reduces miscalibration of all six models --- in all cases our model is closer to the diagonal than the original. We manage to both reduce the under-estimation of the T2T model and the over-confidence of the NMT and GNMT models.

We compare ECE of our method of recalibration to the single temperature method in Table~\ref{tab:ece} (Column ECE). Note the single temperature is selected using the same validation dataset as ours.  Our ECE is lower particularly for the T2T model.   We will show next that our more informed recalibration has several other benefits.

\begin{table}
\begin{center}
\setlength\tabcolsep{3.0pt}
 \begin{tabular}{|l|r r r|r r r| }\hline
 \textbf{Model Name} &  \multicolumn{3}{c|}{\textbf{ECE}} &  \multicolumn{3}{c|}{\textbf{BLEU}}  \\ 
            &   Base & Our & T &   Base & Our & T \\
 \hline\hline
 En-Vi NMT & 9.8  & \textbf{3.5} & 3.8 & 26.2  & \textbf{26.6}  & 26.0 
 \\ \hline
 En-De GNMT4 & 4.8 & \textbf{2.4} & 2.7 & \textbf{26.8} & \textbf{26.8} & 26.7 \\
 \hline
 En-De GNMT8 & 5.0 & 2.2 & \textbf{2.1} & \textbf{27.6} & 27.5 & 27.4  \\
 \hline
 De-En GNMT & 3.3 & \textbf{2.2} & 2.3 & 29.6 & \textbf{29.9}  & 29.6\\ 
 De-En GNMT  &  &  &  & 29.9 & \textbf{30.1}  & \textbf{30.1} \\ 
 (length norm) &  & & & & & \\
 \hline
 De-En NMT & 3.5 & \textbf{2.0} & 2.2 & 28.8 & \textbf{29.0} & 28.7 \\
 \hline
 T2T En-De &  2.9 & \textbf{1.7} & 5.4  & 27.9 & \textbf{28.1}  & 28.1  \\
 T2T En-De(B=4) &  &   &  & \textbf{28.3} & \textbf{28.3} & 28.2 \\
 \hline 
\end{tabular}
\end{center}
\caption{\label{tab:ece} \small Expected Calibration Errors on test data of baseline and models calibrated by two different methods. BLEU is without length normalization, except in De-En GNMT.}
\end{table}

\subsection{An interpretable measure of whole sequence calibration}
For structured outputs like in translation, the whole sequence probability is often quite small and an uninterpretable function of output length and source sentence difficulty.  In general, designing a good calibration measure for structured outputs is challenging. \cite{NguyenO15} propose to circumvent the problem by reducing structured calibration to the calibration of marginal probabilities over single variables.  This works for tractable joint distributions like chain CRFs and HMMs.
For modern NMT systems that assume full dependency, such marginalization is neither tractable nor useful.  We propose an alternative measure of calibration in terms of \bleu\ score rather than structured probabilities.  We define this measure using \bleu\ but any other scoring function including gBLEU, and Jaccard are easily substitutable.

We define model expected $ \bleu_\theta$ of a prediction $\vyp$ as value of \bleu\ if true label sequences were sampled from the predicted distribution $\Pr(\vy|\vx_i,\theta)$ 
\begin{equation}
\begin{split}
 \bleu_\theta(\vyp) &= \text{E}_{\vy \sim P(\vy|\vx_i,\theta)}[{\bleu}(\vyp,\vy)] \\ &\approx \frac{1}{T}\sum_{m=1}^T[{\bleu}(\vyp,\vy_{im})]
\end{split}
\end{equation}
where $\vy_{i1},\ldots,\vy_{iT}$ denote $T$ samples from $P(\vy|\vx_i,\theta)$.\footnote{We could also treat various sequences obtained from beam search with large beam width as samples (unless these are very similar) and adjust the estimator by the importance weights. We observed that both explicit sampling and re-weighted estimates with beam-searched sequences give similar results.}    

It is easy to see that if $P(\vy|\vx_i,\theta)$ is perfectly calibrated the model predicted  $\bleu_\theta$\ will match the actual \bleu\ on the true label sequence $\vy_i$ in expectation. That is, if we considered all predictions with predicted $\bleu_\theta(\vyp)=\alpha$, then the actual \bleu\ over them will be $\alpha$ when $\theta$ is well-calibrated.  This is much like ECE for scalar classification except that instead of matching 0/1 accuracy with confidence, we match actual \bleu\ with expected \bleu. We refer to this as \textit{Structured ECE} in our results (Table \ref{tab:structured_ece}).

\begin{figure*}[ht!]
\begin{center}
\includegraphics[scale=0.48]{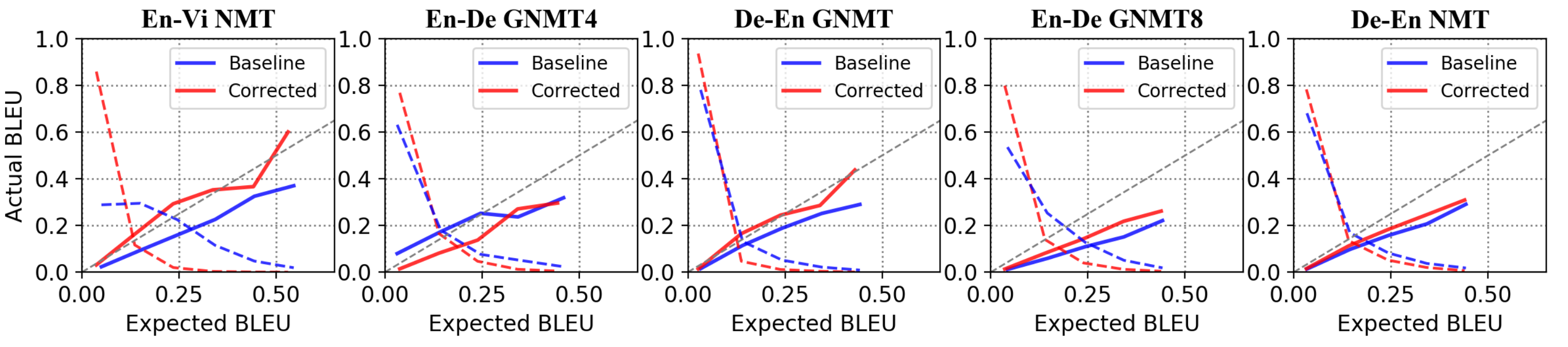}
    \caption{\label{fig:seq} \small Sequence level calibration plots for various models [Baseline + Corrected(Calibrated)]. The dotted lines shows the densities (fraction of all points) in each bin. Note that the density in all the cases shifts to the low end, showing that overestimation is reduced. This trend in the density is same across all models and the calibrated densities are more in agreement with the observed \bleu\ on the datasets (test datasets). }
\end{center}
\end{figure*}

Figure~\ref{fig:seq} shows the binned values of $\bleu_\theta(\vyp)$ (X-axis) and average actual \bleu\ (Y-axis) for WMT + IWSLT tasks on the baseline model and after recalibrating (solid lines).  In the same plot we show the density (fraction of all points) in each bin by each method.  We use $T=100$ samples for estimating $\bleu_\theta(\vyp)$.  Table~\ref{tab:ece} shows aggregated difference over these bins.
We can make a number of observations from these results.  

The calibrated model's \bleu\ plot is closer to the diagonal than baseline's.  Thus, for a calibrated model the $\bleu_\theta(\vyp)$ values provide a interpretable notion of the quality of prediction.  The only exception is the T2T model.  The model has very low entropy on token probabilities and the top 100 sequences are only slight variants of each other, and the samples are roughly identical. An interesting topic for future work is further investigating the reasons behind the T2T model being so sharply peaked compared to other models.




The baseline and calibrated model's densities (shown in dotted) are very different with the calibrated model showing a remarkable shift to the low end. The trend in density is in agreement with the observed BLEU scores, and hence higher density is observed towards the lower end.
 
\begin{table*}
\begin{center}
 \begin{tabular}{|l|r r r|r r r|r r| }\hline
 \textbf{Model Name} &  \multicolumn{3}{c|}{\textbf{ECE}} &  \multicolumn{3}{c|}{\textbf{BLEU}} & \multicolumn{2}{c|}{\textbf{Structured ECE}} \\ 
            &   Base & Our & T &   Base & Our & T & Base & Our\\
 \hline\hline
 En-Vi NMT & 9.8  & \textbf{3.5} & 3.8 & 26.2  & \textbf{26.6}  & 26.0 & 7.3 & \textbf{0.9} 
 \\ \hline
 En-De GNMT4 & 4.8 & \textbf{2.4} & 2.7 & \textbf{26.8} & \textbf{26.8} & 26.7 & 5.8 & \textbf{3.4} \\
 \hline
 En-De GNMT8 & 5.0 & 2.2 & \textbf{2.1} & \textbf{27.6} & 27.5 & 27.4 & 6.4 & \textbf{3.3} \\
 \hline
 De-En GNMT & 3.3 & \textbf{2.2} & 2.3 & 29.6 & \textbf{29.9}  & 29.6 & 2.5 & \textbf{1.3}\\ 
 \hline
 De-En GNMT (Lnorm) & 3.3 & \textbf{2.2} & 2.3 & 29.9 & \textbf{30.1}  & 30.1 & 2.5 & \textbf{1.3}\\ 
 \hline
 De-En NMT & 3.5 & \textbf{2.0} & 2.2 & 28.8 & \textbf{29.0} & 28.7 & 4.0 & \textbf{2.4}\\
 \hline
 T2T En-De &  2.9 & \textbf{1.7} & 5.4  & 27.9 & \textbf{28.1}  & 28.1  & 98.8 & 98.8\\
 \hline
 T2T En-De (B=4) & 2.9 & \textbf{1.7}  & 5.4 & \textbf{28.3} & \textbf{28.3} & 28.2 & 98.8 & 98.8\\
 \hline 
\end{tabular}
\end{center}
\caption{\label{tab:structured_ece} \small Expected Calibration Errors of baseline and models calibrated by two different methods on test set. Structured ECE refers to the ECE from the reliability plot of expected \bleu. We repeat \bleu~and ECE numbers from Table~\ref{tab:ece} for completeness and for easy comparison.}
\end{table*}

\subsection{More Accurate Predictions}
Unlike scalar classification, where temperature scaling
does not change accuracy,
for structured outputs with beam-search like inference, temperature scaling can lead to different MAP solutions.  In Table~\ref{tab:ece} we show the \bleu~score with different methods.  These are with beam size 10, the default in the NMT code.  For the T2T model we report \bleu~with beam size 4, the default in the T2T code. 
In all models except De-En GNMT, using length normalization reduces test\footnote{On the Dev set length normalization improves \bleu, indicating the general difference between the test and dev set in the WMT benchmark} \bleu.  So, we report \bleu~without length norm by default and for De-En GNMT we report with length-norm.  
The table shows that in almost all cases, our informed recalibration improves inference accuracy.  The gain with calibration is more than 0.3 units in \bleu~on three models: En-Vi, De-En GNMT and En-De T2T. Even with length normalization on De-En GNMT, we improve \bleu~by 0.2 using calibration. The increase in accuracy is modest but significant because they came out of only tweaking the token calibration of an existing trained model using a small validation dataset. 

Further calibration using fixed temperature actually {\em hurts accuracy (\bleu).} In five of the six models, the \bleu~after recalibrating with temperature drops, even while the ECE reduction is comparable to ours.  This highlights the importance of accounting for factors like coverage and attention entropy for achieving sound recalibration.



\subsection{BLEU drop with increasing beam-size}
\vspace{-0.1cm}
One idiosyncrasy of modern NMT systems is the drop in \bleu~score as the inference is made more accurate with increasing beam-size.  
In Table~\ref{tab:beam} we show the \bleu\ scores of original models and our calibrated versions with beam size increasing from 10 to 80. These experiments are on the dev set since the calibration was done on the dev set and we want to highlight the importance of calibration. \bleu\ drops much more with the original model than with the calibrated one.  For example for En-De GNMT4, \bleu\ drops from 23.9 to 23.8 to 23.7 to 23.5 as beam width $B$ is increased from 10 to 20 to 40 to 80, whereas after calibration it is more stable going from 23.9 to 23.9 to 23.9 to 23.8.   The \bleu\ drop is reduced but not totally eliminated since we have not achieved perfect calibration.  Length normalization can sometimes help stabilize this drop, but the test accuracy(\bleu) is higher without length normalization on five of the six models. Also,  length normalization is arguably a hack since it is not used during training.  Recalibration is more principled that also provides interpretable scores as a by-product.



{\small
\begin{table}
\begin{center}
 \begin{tabular}{|l|r|r| r| r|} 
 \hline 
 \textbf{Model}     & \textbf{B=10} & \textbf{B=20} & \textbf{B=40} & \textbf{B=80} \\ \hline
 En-Vi NMT & 23.8 & -0.2 & -0.4 & -0.7 \\
        + calibrated   & 24.1 & -0.2 & -0.2 & -0.4 \\ 
 \hline
 En-De GNMT4 & 23.9 & -0.1 & -0.2 & -0.4 \\
  + calibrated & 23.9 & -0.0 & -0.0 & -0.1 \\
 \hline
 En-De GNMT8 & 24.6 & -0.1 & -0.3 & -0.5 \\
   + calibrated &24.7 & -0.1 & -0.4 & -0.6 \\
\hline
De-En GNMT & 28.8 & -0.2 & -0.3 & -0.5 \\
  + calibrated & 28.9 & -0.1 & -0.2 & -0.3 \\
\hline
De-En NMT & 28.0 & -0.1 & -0.4 & -0.6 \\
  + calibrated & 28.2 & -0.0 & -0.2 & -0.2\\
\hline
En-De T2T* & 26.5 & -0.2 & -0.7 & -1.2 \\
  + calibrated & 26.6 & -0.1 & -0.3 & -0.4\\
\hline
\end{tabular}
\end{center}
\caption{\label{tab:beam} \small BLEU with increasing beam on the devset. *Beam sizes for Transformer/T2T: 4, 8, 10 and 12}
\vspace{-1em}
\end{table}
}

\section{Related Work}
Calibration of scalar classification and regression models has been extensively studied. \citet{05predicting} systematically evaluated many classical models and found models trained on conditional likelihood like logistic regression and neural networks (of 2005) to be well-calibrated, whereas SVMs and naive Bayes were poorly calibrated. \citet{NguyenO15} corroborated this for NLP tasks.  Many methods are proposed for fixing calibration including Platt's scaling~\cite{Platt99}, Isotonic regression~\cite{Zadrozny2002}, and Bayesian binning~\cite{Naeini2015},  and training regularizers like MMCE~\cite{kumar2018}.
A principled option is to capture parameter uncertainty using Bayesian methods. Recently, these have been applied on DNNs using variational methods~\cite{louizos17a}, ensemble methods~\cite{LakshmiN17}, and weight perturbation-based training~ \cite{KhanNTLGS18}.

For modern neural networks, a recent systematic study~\cite{GuoPSW17} finds them to be poorly calibrated and finds temperature scaling to provide the best fix.
%
We find that temperature scaling is inadequate for more complicated structured models where different tokens have very different dynamics.  We propose a more precise fix derived after a detailed investigation of the root cause for the lack of calibration.
 
Going from scalar to structured outputs, \citet{NguyenO15} investigates calibration for NLP tasks like NER and CoRef on log-linear structured models like CRFs and HMMs. They define calibration on token-level and edge-level marginal probabilities of the model.   \citet{Kuleshov15} generalizes this to structured predictions. But these techniques do  not apply to modern NMT networks since each node's probability is conditioned on all previous tokens making node-level marginals both intractable and useless.  

Concurrently with our work, \citet{ott18a} studied the uncertainty of neural translation models where their main conclusion was that existing models "spread
too much probability mass across sequences".  However, they do not provide any fix to the problem. Another concern is that their observations are only based on the FairSeq's CNN-based model, whereas we experiment on a much larger set of architectures. Our initial measurements on a pre-trained En-Fr FairSeq model\footnote{Downloaded from \texttt{https://github.com/pytorch/} \texttt{fairseq}, commit \texttt{f6ac1aecb3329d2cbf3f1f17106b74} \texttt{ac51971e8a}.} found the model to be well-calibrated (also corroborated in their paper) unlike the six architectures we present here (which they did not evaluate).  An interesting area of future work is to explore the reasons for this difference.

The problem of drop in accuracy with increasing beam-size and length bias has long puzzled researchers~\cite{BahdanauCB14,Sountsov16,KoehnK17} and many heuristic fixes have been proposed including the popular length normalization/coverage penalty~\cite{Wu16}, word reward~\cite{He2016ImprovedNM}, and bounded penalty~\cite{Yang2018BreakingTB}. These heuristics fix the symptoms by delaying the placement of the EOS token, whereas ours is the first paper that attributes this phenomenon to the lack of calibration.  Our experiments showed that miscalibration is most severe for the EOS token, but it affects several other tokens too.  Also, by fixing calibration we get more useful output probabilities, which is not possible by these fixes to only the BLEU drop problem. 


\section{Conclusion and Future work}
Calibration is an important property to improve interpretability and reduce bias in any prediction model. For sequence prediction it is  additionally important for sound functioning of beam-search or any approximate inference method. We measured the calibration of six state-of-the-art neural machine translation systems built on attention-based encoder-decoder models using our proposed weighted ECE measure to quantify calibration of an entire multinomial distribution and not just the highest confidence token.

The token probabilities of all six NMT models were found to be surprisingly miscalibrated even when conditioned on true previous tokens. On digging into the reasons, we found the EOS token to be the worst calibrated.  Also, positions with higher attention entropy had worse calibration. 

We designed a parametric model to recalibrate as a function of input coverage, attention uncertainty, and token probability. We achieve significant reduction in ECE and show that 
%
%
translation accuracy improves by as much as 0.4 when the right models are used to fix calibration.  Existing temperature scaling recalibration actually worsens accuracy. We show that improved calibration leads to greater correlation between probability and error and this manisfests as reduced \bleu\ drop with increasing beam-size. 
We further show that in our calibrated models the predicted \bleu\ is closer to the actual \bleu.

We have reduced, but not totally eliminated the miscalibration of modern NMT models. Perhaps the next round of fixes will emerge out of a better training mechanism that achieves calibration at the time of training. 
The insights we have obtained about the relation between coverage and EOS calibration, and attention uncertainty should also be useful for better training. 
\vspace{1cm} 
\section*{Acknowledgements}
We thank all anonymous reviewers for their comments. We thank members of our group at IIT
Bombay, at the time when this research was being carried out, for discussions. Sunita Sarawagi gratefully acknowledges the support of NVIDIA corporation for Titan X GPUs. 

\clearpage

\bibliographystyle{acl_natbib}
\bibliography{MLCopied,pubs2}

\end{document}